\documentclass[sigconf]{acmart}
\AtBeginDocument{%
  \providecommand\BibTeX{{%
    \normalfont B\kern-0.5em{\scshape i\kern-0.25em b}\kern-0.8em\TeX}}}

\acmBooktitle{Woodstock '18: ACM Symposium on Neural Gaze Detection,
 Oct. 29--Nov. 3, 2023, Ontario, Canada} 
\acmPrice{15.00}
\acmISBN{978-1-4503-XXXX-X/18/06}

\acmSubmissionID{2578}

\usepackage{multirow}
\usepackage{colortbl}
\usepackage{booktabs}

\begin{document}

% \title{Multi-scale Radiance Field Reconstruction from Multi-View Stereo with Depth Guidance and Feature Fusion}
% \title{DGNF-NeRF:Depth Guidance and Neighbors Fusion of Multi-scale Neural Radiance Fields}
\title{Improved Neural Radiance Fields Using Pseudo-depth and Fusion}


\author{Jingliang Li, Qiang Zhou, Chaohui Yu, Zhengda Lu, Jun Xiao, Zhibin Wang, Fan Wang}

\begin{abstract}
Since the advent of Neural Radiance Fields, novel view synthesis has received tremendous attention. The existing approach for the generalization of radiance field reconstruction primarily constructs an encoding volume from nearby source images as additional inputs. 
However, these approaches cannot efficiently encode the geometric information of real scenes with various scale objects/structures.
In this work, we propose constructing multi-scale encoding volumes and providing multi-scale geometry information to NeRF models.
To make the constructed volumes as close as possible to the surfaces of objects in the scene and the rendered depth more accurate, we propose to perform depth prediction and radiance field reconstruction simultaneously. The predicted depth map will be used to supervise the rendered depth, narrow the depth range, and guide points sampling.
Finally, the geometric information contained in point volume features may be inaccurate due to occlusion, lighting, etc. To this end, we propose enhancing the point volume feature from depth-guided neighbor feature fusion.
Experiments demonstrate the superior performance of our method in both novel view synthesis and dense geometry modeling without per-scene optimization.
\end{abstract}







\maketitle

\section{Introduction}

Novel view synthesis is a fundamental problem for computer vision community, which aims to produce photo-realistic images of the same scene at novel viewpoints. This long-standing problem has recently received tremendous attention \cite{liu2020neural} due to the advent of neural rendering.
Notably, the recent method of neural radiance fields (NeRF) \cite{mildenhall2021nerf} has shown impressive performance on novel view synthesis, represented using global MLPs. Although NeRF and its extensions \cite{martin2021nerfw,zhang2020nerf++} can represent scenes faithfully and compactly, they typically require a very long per-scene optimization process to obtain high-quality radiance fields.
Recently, some multi-view rendering literature~\cite{yu2021pixelnerf,wang2021ibrnet,chen2021mvsnerf,xu2022point,lin2022efficient} have been proposed to address these shortcomings, which can generalize well across scenes by taking advantage of the nearby input views. These methods construct encoding volumes from source images and extract volume features for each sample point via trilinear interpolation. The point volume feature contains the geometric information of the scene, which will be used as an additional input to the NeRF~\cite{mildenhall2021nerf} model to improve the model's scene generalization ability.

\begin{figure}[t]
  \centering
  \includegraphics[width=0.96\linewidth]{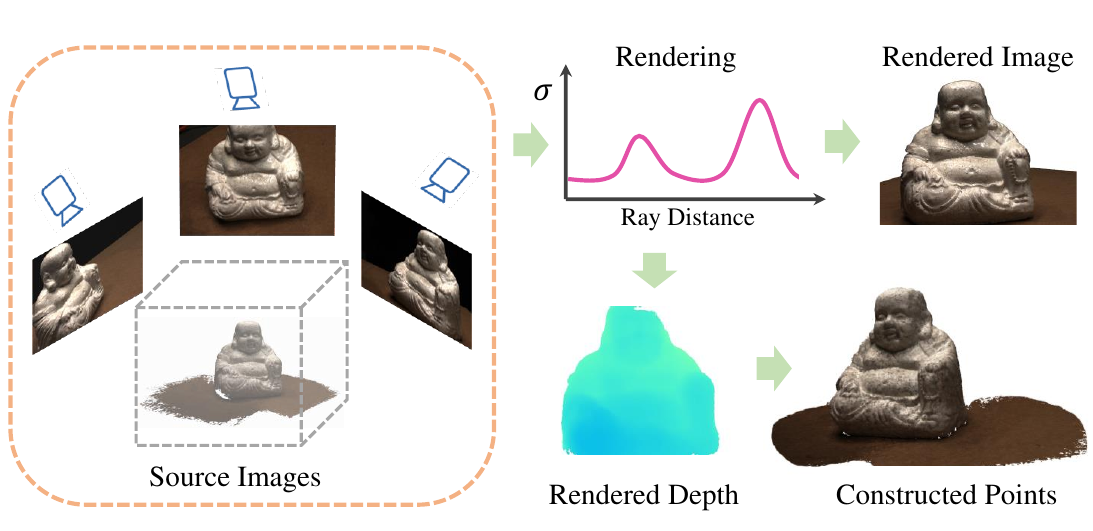}
  \vspace{-.1in}
  \caption{We present a more effective end-to-end method for both photo-realistic novel views rendering and high-quality dense geometry modeling. Our method utilizes multi-scale radiance field, auxiliary depth prediction head, and feature fusion for more realistic novel views, accurate depth maps, and dense points.
  }
  \label{fig:teaser}
  \vspace{-.1in}
\end{figure}

% 总述一下，还方便引入Figure 1
Although these methods achieve promising results, they still have several limitations, especially when generalizing to scenes with diverse objects/structures and modeling delicate geometry. In this work, as depicted in Figure~\ref{fig:teaser}, we present an effective end-to-end method for both photo-realistic novel views rendering and high-quality geometry modeling.

\begin{figure*}
  \includegraphics[width=0.98\textwidth]{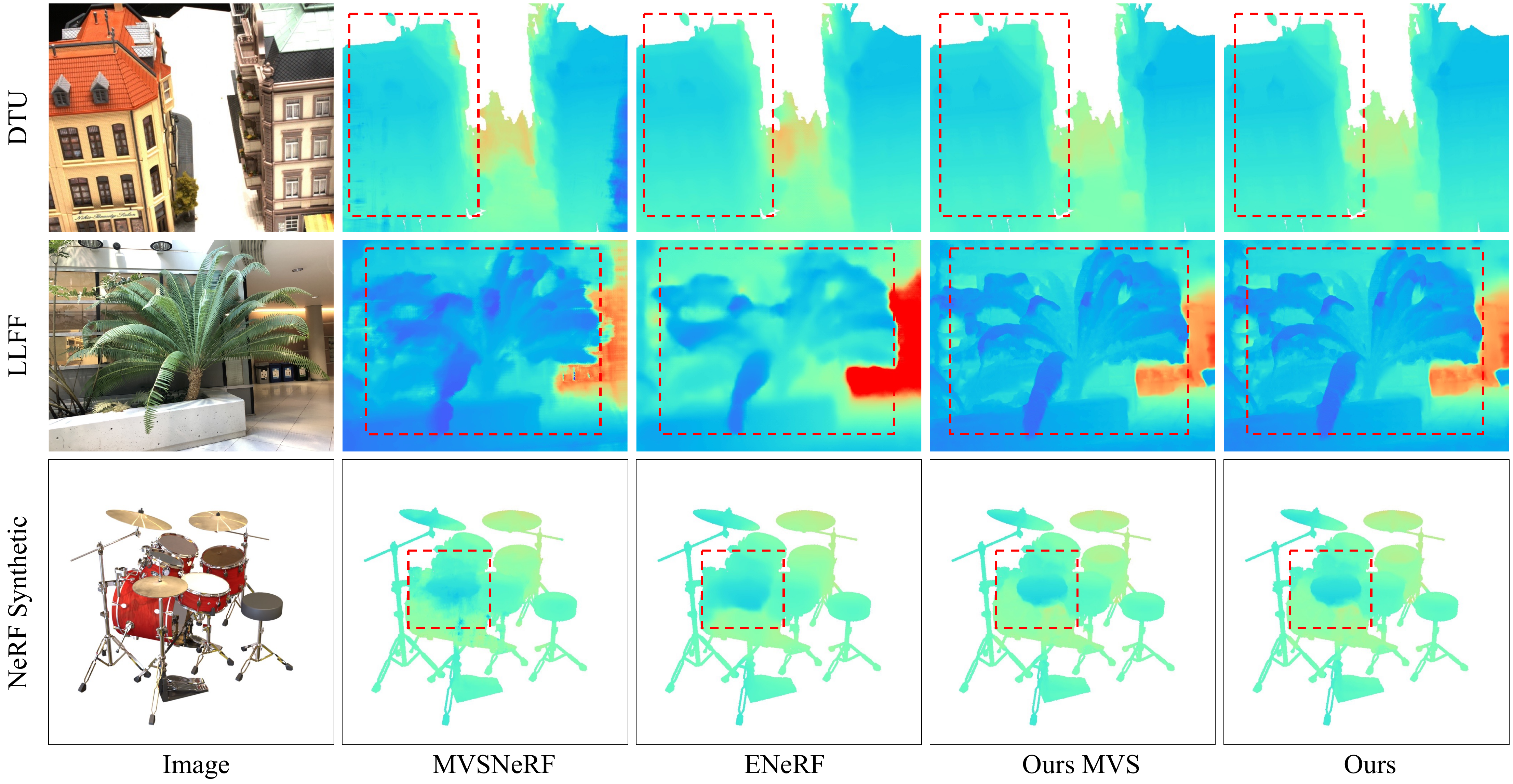}
  %\caption{Visual illustrations of synthetic source depths and rendered reference images. The black region in synthetic depth represent the uncertainty region due to occlusion.}
  \vspace{-.1in}
  \caption{Visual illustrations of rendered source depths of different methods. Our method can learn more delicate geometry for neural radiance fields. ``Ours MVS'' means the output of the auxiliary depth prediction head.} 
  \label{fig:1_depth_img}
\end{figure*}

% 要么采用总分结构，直接说明现存challenge，提出我们方法，并在这里提一下Figure 1
% 要么最introduction的最后，总结的时候提Figure 1，说明我们框架的特点。

% 1. why using multi-scale
The size of objects/structures in a realistic scene is diverse, and thus a single scale encoding volume is not adequate to provide geometric information for the entire scene. To this end, inspired by recent coarse-to-fine multi-view stereo (MVS) approaches, we propose constructing pyramid-structured encoding volumes to provide geometric information at all scales.
The low-resolution encoding volume provides more geometric information for large objects/structures in the scene. In contrast, the high-resolution encoding volume focuses more on small-scale objects/structures.
For each scale encoding volume, we sample points on the ray, trilinear interpolate point volume features and regress the volume density and view-dependent radiance in the scene.
That is, we reconstruct the multi-scale radiance fields and combine the reconstructed radiance fields at all scales to perform the final volume rendering for view synthesis.
%That is, we reconstruct multi-scale radiance fields. Finally, we combine the reconstructed radiance fields at all scales and perform final volume rendering for view synthesis.

% 2. why using depth
Building pyramid encoding volumes without depth prior guidance is a massive challenge for GPU memory usage and rendering rate, since we need to sample enough points across the entire depth range when building each scale volume.
Although using RGB values to supervise the model helps to create more realistic novel views for certain scenes, there may still be challenges with ambiguous geometric information, particularly in texture-less scenes.
Recently, some works~\cite{roessle2022dense,deng2022depth} have proposed depth-guided sampling for NeRF to reduce the number of sampling points, where depth information comes from third-party software or models, such as COLMAP~\cite{schoenberger2016sfm} and depth completion models.
In this work, we propose to train an additional depth prediction head along with the radiance field reconstruction, and use the predicted depth map of the depth head to supervise the rendered depth, reduce the depth range and guide point sampling.
The depth head is trained in an unsupervised manner with the same loss function as~\cite{xu2021jdacs}.
As shown in Figure~\ref{fig:1_depth_img}, our method using pseudo-depth guidance can learn more compact geometry compared to other counterparts.

% 3. why using featrue fusing
Point volume features trilinear interpolated from encoding volumes play an essential role in rendering precision across scenes. 
However, the geometric information contained in point volume features may be inaccurate due to occlusion, lighting, etc. Aggregating the volume features of nearby points from the same surface can effectively improve the correctness of the geometric information in point volume features.
To this end, we propose a depth-guided adaptive neighbor feature fusion module. Utilizing the predicted depth map, we adaptively select adjacent points from the same surface and fuse these features using a shared multi-head cross-attention module.
The projected features of these adaptive neighbor points on the source view are also input into NeRF as auxiliary information.

Our approach is fully differentiable and can therefore be trained end-to-end using multi-view images.
Similar to CasMVSNet \cite{gu2020cascade}, we build multi-scale encoding volumes at the target view by warping 2D image features from source views onto sweeping planes in the target view. 
The multi-scale encoding volumes are used for depth maps prediction and radiance fields (including density $\sigma$ and view-dependent radiance $r$) reconstruction at the same time.
Finally, we combine the reconstructed multi-scale radiance fields and perform view synthesis via differentiable ray marching.
In summary, our contributions are:

\begin{itemize}
%感觉可以先一句话总结框架贡献
\item We present a more efficient multi-scale framework for improving NeRF, which is capable of rendering photo-realistic novel views as well as high-quality dense geometry.
\item We propose an auxiliary depth maps prediction head, used for supervising rendered depth, narrowing the depth range and guide points sampling.
\item We propose enhancing the point volume feature, warped feature and color information from depth-guided neighbor feature fusion.
\item Extensive experiments show that our framework achieves state-of-the-art results on various view synthesis datasets.
\end{itemize}

\section{Related Work}

\paragraph{\textbf{View synthesis.}} 
The long-standing problem of novel view synthesis is fundamental in computer vision. Recently, various neural scene representations have been presented \cite{bi2020deep,lombardi2019neural,sitzmann2019deepvoxels,zhou2018stereo}, due to the advent of neural rendering.
NeRF \cite{mildenhall2021nerf} has shown impressive performance, which uses global MLPs to regress the volume density and view-dependent radiance at any arbitrary point in the space and applies volume rendering to synthesize images at novel viewpoints. Following works have highlighted different tasks such as composing and editing \cite{xu2022deforming,wu2022object,yuan2022nerf,kania2022conerf}, large-scale scenes synthesizing \cite{tancik2022block,turki2022mega}, relighting \cite{srinivasan2021nerv,bi2020neural,boss2021nerd}. 
Like NeRF, most of these works construct radiance fields using global MLPs for the entire space, which requires a lengthy optimization process for each new scene before they can synthesize any novel views of that scene.
To achieve cross-scene estimation for view synthesis, borrowing from the deep multi-view stereo, several approaches improve the cross-scene generalization ability of models by encoding scene geometry information.

\paragraph{\textbf{Multi-view stereo.}} Multi-view stereo (MVS) is a classical computer vision problem, which aims to reconstruct the geometry from multiple viewpoints. Recently, learning-based multi-view stereo methods have been introduced and achieve impressive results. MVSNet \cite{yao2018mvsnet} first leveraged the plane sweep-based cost volume formulation followed by 3D CNN for regularization to predict the depth maps. However, methods relying on 3D cost volumes are often limited to low-resolution input images, because 3D CNNs are generally time and GPU memory-consuming. Following works have extended this technique with recurrent plane sweeping \cite{yao2019recurrent}, confidence-based aggregation \cite{luo2019p} and multi-scale cost volumes \cite{gu2020cascade,wang2021patchmatchnet}, improving the efficiency and reconstruction quality.
In a coarse-to-fine fashion, early cost volumes are built on coarser-resolution features with sparsely sampled depth hypotheses, which result in relatively low volumetric resolution. Subsequently, estimated depth maps from earlier stages are used to narrow the depth range and construct thinner cost volumes on finer-resolution features.
Drawing on these works, we propose to train depth prediction and radiance field reconstruction simultaneously. With the predicted depth, we construct efficient pyramid encoding volumes that provide multi-scale geometric information for the NeRF module.

\paragraph{\textbf{NeRF based on multi-view.}} Although MVS methods can reconstruct the geometry of the scene, they are sensitive to potential inaccuracies in point clouds from corrupted depth, especially when there are thin structures and textureless regions. In contrast, NeRF-based methods model scenes as neural volumetric radiance fields and can reproduce the faithful
scene appearance, producing photo-realistic novel views.
To improve the generalization of NeRF, some multi-view rendering methods are proposed.  PixelNeRF \cite{yu2021pixelnerf} takes spatial image features aligned to each pixel as an input and learns a scene prior for reconstructing. IBRNet \cite{wang2021ibrnet} generates a continuous scene radiance field on-the-fly from multiple source views for rendering novel views. 
MVSNeRF~\cite{chen2021mvsnerf} and ENeRF~\cite{lin2022efficient} utilize a plane swept 3D cost volume for geometric-aware scene understanding, using only a few images as input.
Point-NeRF \cite{xu2022point} models a volumetric radiance field with a neural point cloud using multi-view images.
These works use a single-scale encoding volume, which is insufficient to provide adequate geometric information for scenes with objects of various scales.
In this work, we propose constructing pyramid-structured encoding volumes to provide geometric information at all scales.

\section{Approach}

\begin{figure*}[t]
  \centering
  \includegraphics[width=0.98\textwidth]{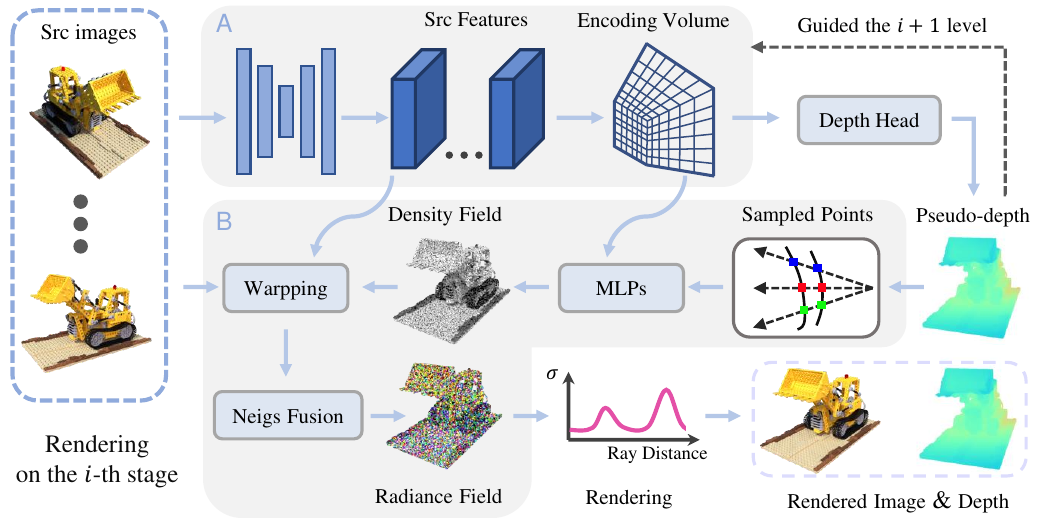}
  \caption{Overview of our pipeline, \textit{A} represents the encoding volume reconstruction, while \textit{B} represents the neural field reconstruction. Our framework first identify a set of neighboring source views and extract hierarchical feature maps. Then, we apply inverse warping, 3D CNN, and auxiliary depth heads to reconstruct encoding volumes and estimate depth maps at each stage. 
  Afterward, the multi-scale neural fields (including density field and radiance field) are reconstructed by trilinear interpolation and MLPs.
  Finally, we apply depth-guided adaptive feature fusion to enhance the radiance field obtained by fusing neighbors features.
  Differentiable ray marching uses multi-scale neural fields for final rendering.
  }
  \label{fig:pipeline}
\end{figure*}

Our framework has the same input settings as MVSNeRF~\cite{chen2021mvsnerf}, i.e., additional $N$ source images and their camera parameters are provided when performing view synthesis.
As shown in Fig.~\ref{fig:pipeline}, 
our framework constructs pyramid neural encoding volumes to provide multi-scale geometric information for the NeRF module (Sec.~\ref{sec:multi-scale-radiance}).
The multi-scale encoding volumes will be utilized to regress the depth maps and reconstruct the multi-scale radiance fields.
The key to efficient pyramid volume construction is effectively reducing the depth range. To this end, we propose to train depth prediction and radiance field reconstruction simultaneously (Sec.~\ref{sec:aux-depth-head}).
The additional input $f^l$ to the NeRF module in Eq.~\ref{eq:level-regress} is trilinear interpolated from the $l$-th level encoding volume $V^l$.
To make the geometric information in the point features more robust, we further explore enhancing the point volume feature $f^l$ by fusing multi-scale volume features, and adaptive neighbor volume features, respectively (Sec.~\ref{sec:feat_fusion}).

\subsection{Multi-scale neural field reconstruction} % add by zhou
\label{sec:multi-scale-radiance}

In this section, we introduce our multi-scale neural field reconstruction framework.
Our framework learns an encoding volume and reconstructs the density field and radiance field separately.
During training, the pixel is rendered at each level to facilitate the acquisition of multi-scale geometric information. Conversely, during testing, the pixel is solely rendered at the final level.

\paragraph{\textbf{Feature pyramid.}}
Similar to MVSNet~\cite{yao2018mvsnet}, a shared eight-layer 2D CNN is applied to extract deep features for all source images $\{I_i\}_{i=1}^{N}$, $I_i\in{\mathbb{R}^{H\times W}}$. Afterward, we use a feature pyramid network~\cite{lin2017feature} to extract hierarchical feature maps ${F^1, \dots, F^L}$ with different spatial resolutions, where $L$ denotes the number of levels. These hierarchical feature maps will be used to construct multi-scale encoding volumes.

\paragraph{\textbf{Encoding volume.}}
Taking the $l$-th level as an example. For each pixel $p$ in the reference view, we sample $|\mathcal{S}_\text{A}^l|$ points, where $\mathcal{S}_\text{A}^l$ denotes the $l$-th level sampling points for encoding volume, within the depth range $\mathcal{R}^l$ (note that the depth range can be different for different pixels) 
and build encoding features for each sampled point. Following learning-based methods~\cite{yao2018mvsnet}, we construct encoding volume by warping the source features into reference views.
Specifically, for a sample point with pixel coordinate $p$ and depth value $d$,
we first warp it to the $i$-th source view via inverse warping:
\begin{equation}
    p_i = K_i T_{j} (d \cdot K^{-1} p),
    \label{eq:inverse_warping}
\end{equation}
where $K$ and $K_i$ are the intrinsics for the reference and the $i$-th source image, respectively.
$T_i$ is the relative transformation from reference to $i$-th source image.
$p_i$ is the warped pixel location in the $i$-th source image.
Based on the inverse warping, we construct the encoding volume $\bar{V}^l \in \mathbb{R}^{|\mathcal{S}_r^l| \times \frac{H}{2^{L-l}} \times \frac{W}{2^{L-l}} \times C}$ by computing the variance of multi-view source features for each point.
\begin{equation}
    \bar{V}^l = \text{Var} (F^l_1[p_1], \dots, F^l_N[p_N]),
\end{equation}
where $N$ denotes the number of source images, 
$F^l_N[p_N]$ represents the bilinearly interpolated feature at the $l$-th level feature map of the $N$-th source image, using the warped pixel coordinate $p_N$.

To smooth the noise contaminated in the raw encoding volume, we apply a deep 3D U-Net to regularize the encoding volume and build a new encoding volume, and represent the similarity of the source features on different depth values. This process is expressed by:
\begin{equation}
    %V^l = B(\bar{V}^l), \text{UNet}_{3D}(\bar{V}^l).
    V^l = \text{UNet}_{\text{3D}}(\bar{V}^l).
\end{equation}

\paragraph{\textbf{Density field reconstruction.}}
The higher similarity, the greater the probability that the point is located on the surface of the object, which means that the encoding volume contains the density information of the scene.
Existing methods construct the density field using the 3D position and viewing direction of the point, which lacks guidance from global and local information and leads to discontinuities in the rendered depth map.
Different from these methods, we reconstruct the density by sharing the encoding volume, with a tiny MLP network.

Specifically, along a ray passing through the pixel $p$ of the $l$-th level, we sampled a set of points $\mathcal{S}_\text{B}^l$ for neural field reconstruction. For each point $s_b \in \mathcal{S}_\text{B}^l$, we regress the point's density $\sigma$ using a tiny MLP network $\text{MLP}_1$, which only contains two layers MLP. The first layer is used for fusing the feature volume and the direction information, and the second layer aims to regress the density:
\begin{equation}
    \sigma,\ f_h = \text{MLP}_1(f_v, \ \bigtriangleup d),
\end{equation}
where $f_h$ is the hidden features, and $f_v$ is the volume feature, obtained by trilinear interpolation in the $l$-th level encoding volume $V^l$ at position $x$, 3d position of the point $s_b$. $\bigtriangleup d$ is defined as the concatenation of the direction of $d_i - d$ from all source views, $d$ and $d_i$ are the ray directions of the point under the reference view and corresponding source view.
\begin{equation}
    \bigtriangleup d = Concat(d_1 - d, \dots, d_N - d).
\end{equation}

\paragraph{\textbf{Radiance field reconstruction.}}
Same with density field reconstruction, we also perform volume attribute regression at each level. Following with IBRNet~\cite{wang2021ibrnet}, we regress the point's $s_b$ radiance, viewed in direction $d$, by predicting blending weights for the image colors $\{c_i\}_{i=1}^N$ in the source views. The weight of the $i$-th source view is predicted by an MLP network $\text{MLP}_2$ :
\begin{equation}
    w_i = \text{MLP}_2(\bigtriangleup d,\ f_h,\ c_i,\ f_i),
\label{eq:level-regress}
\end{equation}
where $c_i$, $f_i$ are the RGB value and the image features of the pixel at the projected position of location  $x$ in the $i$-th source view. The color of the point $s_r$ location in the direction $d$ is blended via a soft-argmax operator, as the following:
\begin{equation}
    r = \sum_{i=1}^{N}\frac{exp(w_i)c_i}{\sum\nolimits_{j=1}^N exp(w_i)}.
\end{equation}

\paragraph{\textbf{Volume rendering.}}
To render the pixel color and density, we perform the differentiable ray marching~\cite{mildenhall2021nerf} based on the reconstructed radiance fields $\{\sigma^l, r^l\}_{l=1}^L$. During testing, we only perform the rendering at the final level. In contrast, we conduct rendering at each level during the training. By supervising the color and density at each level, the model can more effectively learn multi-scale information, thereby improving its ability to generalize.

Specifically, for the $l$-th level, the color $\tilde{c}_l$ and depth $\tilde{d}_l$ of the pixel $p$ is calculated by accumulating the radiance of sampling points at each level, and is given by:
\begin{equation}
\begin{split}
    &\tilde{c}_l = \sum_{k=1}^{|\mathcal{S}_B^l|}\tau_k(1 - \text{exp}(-\sigma_k))r_k, \text{\quad} \tilde{d}_l = \sum_{k=1}^{|\mathcal{S}_B^l|}{\sigma_k}{z_k}
\end{split}
\label{eq:volume_rendering_2}
\end{equation}
\begin{equation}
    \text{where \quad} \tau_k = \text{exp}(-\sum_{j=1}^{k-1}\sigma_j),
\end{equation}
and $z_k$ is the depth value of the point $s_b \in \mathcal{S}_B^l$ in the reference view. And $\tau$ represents the volume transmittance.

\subsection{Auxiliary depth prediction head} % add by zhou
\label{sec:aux-depth-head}

Due to the lack of depth guidance, MVSNeRF sample points uniformly across the entire depth range for each pixel. The number of sample points is a trade-off between rendering precision and rate.
Inspired by CasMVSNet~\cite{gu2020cascade}, we append a depth prediction head $F_{\text{depth}}$ (a convolutional layer) after each scale encoding volume, and predict the depth map $D^l$ with confidence $U^l$ in the target view:
\begin{equation}
    D^l, U^l = F_\text{depth} (V^l).
\end{equation}

Based on the encoding volume, the core of the depth prediction head compresses the similarity of source features. As a result, the predicted depth can have higher accuracy compared to the regression approach~\cite{yao2018mvsnet}, particularly in regions with abundant texture information.

Utilizing the predicted depth map, We first supervise the pixel depth calculated by volume rendering, as described in Sec~\ref{training}. Besides, we can reduce the number of sampling points for radiance field reconstruction, and narrow down the depth range of the next level, thus resulting in better rendering precision with fewer sample points:
\begin{equation}
\begin{split}
    &|\mathcal{S}_B^l| = |\mathcal{S}_A^l| * \alpha_l, \\
    &\mathcal{R}^{l+1} = \mathcal{R}^l * \beta_l,
\end{split}
\end{equation}
As described in Sec.~\ref{sec:multi-scale-radiance}, $\mathcal{S}_\text{A}^l$ and $\mathcal{S}_\text{B}^l$ denote the $l$-th level sampling points for encoding volume and radiance field reconstruction, respectively.  %SA SB 最好第一次提到的时候就解释，前面看到的时候reviewer可能不知道代表什么
Points $\mathcal{S}_A^l$, with $N_A^l=|\mathcal{S}_A^l|$ sampled points, are uniformly sampled in the depth range $\mathcal{R}^l$ for the $l$-th level, and the reducing factor $\alpha_l (<1)$ is used to decrease the number of points for radiance field reconstruction.
Specifically, for $\mathcal{S}_B^l$, we select $N_B^l=|\mathcal{S}_A^l| * \alpha_l$ points closest to the predicted depth from $\mathcal{S}_A^l$.
In addition, we narrow the depth range of the next level $\mathcal{R}^{l+1}$ to $\mathcal{R}^l * \beta_l$ centered on the predicted depth value, in which $\beta_l <1$ is the reducing factor.
By default, we use three levels and set [$N_A^l$], [$\alpha_l$], [$\beta_l$] to [48, 32, 8], [1/6, 1/4, 1/2] and [1/6, 1/16] respectively.

\subsection{Depth-guided adaptive feature fusion} \label{sec:feat_fusion}

The geometric information contained in the point volume feature $f^l$ may be inaccurate due to
occlusion, lighting, etc. This section introduces our feature fusion strategies to enhance the point volume features.

The MVS methods~\cite{wang2021patchmatchnet,xu2020aanet} adaptively aggregate cost volume features to improve matching robustness, using reference view images as input to predict adjacent pixel offsets.
However, reference view images are not available in the view synthesis task. 
To this end, we propose to use the predicted depth map $D^l \in \mathbb{R}^{H \times W \times 1}$ to adaptively enhance the hidden volume features $f^h$, projected colors $c_i$ and projected features $f_i$ in Eq.~\ref{eq:level-regress}.

For the adaptive fusion of point volume features $f_v$ in Eq.~\ref{eq:level-regress}, we employ the depth map and inverse warped source image features as inputs to predict the pixel offsets.
Specifically, we first obtain the warped features in the reference view by inverse warping, taking the predicted depth map and source image features as input:
\begin{equation}
    \hat{F}^l = Concat(IW(F_1^l,\ D^l),\dots,IW(F_N^l,\ D^l)),
\end{equation}
where $F_i$ denotes the features of the $i$-th source image, $N$ denotes the number of source images, $D^L$ is the predicted depth map in the reference view, $IW(\cdot)$ denotes the inverse warping in Eq.~\ref{eq:inverse_warping}, and $\hat{F}^l$ is the warped image features in the reference view. 
Then, we concatenate the depth map and warped features and apply a 2D CNN to predict the pixel offsets $\{\Delta p_k\}_{k=1}^{K}$ of the pixel $p$.
The final neighbor volume feature fusion is obtained using a single-layer multi-head cross-attention module $F_{\text{MHCA}}$, taking the volume feature at location $p$ as the query and the volume features at locations $\{p + \Delta p_k\}_{k=1}^{K}$ as keys and values:
\begin{equation}
    \widetilde{f}_v = F_{\text{MHCA}}(\ V(p),\; \{V(p + \Delta p_k)\}_{k=1}^{K}\;).
\end{equation}

\begin{table*}[!t]
  \centering
  \caption{Quantitative results of novel view synthesis.
  We show averaged results of PSNRs, SSIMs, and LPIPs on three different datasets.
  For the Realistic Synthetic NeRF dataset, the two numbers in each item refer to the evaluation of the central/entire regions of the novel images.}
  \begin{tabular}{
p{2.cm}
p{1.5cm}<{\centering}p{1.5cm}<{\centering}p{1.5cm}<{\centering}
p{0.2cm}
p{1.0cm}<{\centering}p{1.0cm}<{\centering}p{1.0cm}<{\centering}
p{0.2cm}
p{1.0cm}<{\centering}p{1.0cm}<{\centering}p{1.0cm}<{\centering}
}
    \toprule
    \multirow{2}{*}{Method}
    & \multicolumn{3}{c}{NeRF Synthetic}
    & &  \multicolumn{3}{c}{DTU}
    & &  \multicolumn{3}{c}{Real Forward-Facing} \\
    \cmidrule{2-4} \cmidrule{6-8} \cmidrule{10-12}
    & PSNR $\uparrow$ & SSIM $\uparrow$ & LPIPS $\downarrow$
    & & PSNR $\uparrow$ & SSIM $\uparrow$ & LPIPS $\downarrow$
    & & PSNR $\uparrow$ & SSIM $\uparrow$ & LPIPS $\downarrow$ \\
    \midrule
    PixelNeRF &  -/7.39 & -/0.658 & -/0.411 & & 19.31 & 0.789 & 0.382 & & 11.24 & 0.486 & 0.671 \\
    IBRNet    & 21.91/22.44 & 0.857/0.874 & 0.203/0.195 & & 26.04 & 0.971 & 0.191 & & 21.79 & 0.786 & 0.279 \\
    MVSNeRF   & 23.62/24.63 & 0.897/0.929 & 0.176/0.155 & & 26.63 & 0.931 & 0.168 & & 21.93 & 0.795 & 0.252 \\
    ENeRF     & 23.25/26.65   & 0.893/0.947 & 0.152/0.072   & & 27.61 & 0.956 & 0.091 & & 22.78 & 0.808 & 0.209 \\
    Ours      & 24.88/26.76 & 0.902/0.950 & 0.110/0.067 & & 28.52 & 0.952 & 0.110 & & 22.78 & 0.781 & 0.243 \\
    \bottomrule
\end{tabular}
  \label{tab:res_color}
\end{table*}

For projected colors and features $c_i$, $f_i$ in Eq.~\ref{eq:level-regress}, we employ the same adaptive fusion approach, taking adaptive colors $c_i$ fusion as an example.
Specifically, for each predicted neighborhood pixel $\hat{p}_k = p + \Delta p_k$, we project it to the corresponding source images and obtain the project color value $c_{k,i}$:
\begin{equation}
    c_{k,i} = I_i\left[ p_{k, i}\right],
\end{equation}
where $p_{k, i}$ is the projected pixel location in the $i$-th source image $I_i$.
The final colors input to the NeRF module is the attention warped colors of predicted neighbors, which can be formulated as:
\begin{equation}
    \widetilde{c}_i = F_\text{MHCA}(c_i,\ \{c_{k,i}\}_{k=1}^{K}).
\end{equation}

\subsection{End-to-end training} \label{training}

In particular, we train our full pipeline on the DTU dataset~\cite{aanaes2016large}, which can learn a powerful generalizable function, reconstructing radiance fields across scenes from only three input nearby source images.

For the auxiliary depth prediction, we adopt the same loss function as in unsupervised MVS tasks~\cite{khot2019learning} to improve the accuracy of predicted depth. This, in turn, helps more effectively supervise the depth of the rendering branch, guide the sampling points, and adjust the depth range. The loss is defined as follows:
\begin{equation}
    L_{mvs} = L_{PC} + L_{SSIM} + L_{Smooth},
\end{equation}
where, $L_{PC}$, $L_{SSIM}$, and $L_{Smooth}$ are the photometric consistency loss, structured similarity loss, and depth smoothness loss. Please refer to ~\cite{khot2019learning} for more details.

We optimize the outputs of our rendering approach, including the color $\tilde{c}$ and depth $\tilde{d}$, with the ground truth target view $I$ and the pseudo depth $D$.
The auxiliary depth prediction head predicts the pseudo-depth based on the similarity of source features. Although it may have higher accuracy in texture-rich regions, it may be unreliable in texture-less regions. Therefore, we only consider pixels with a confidence score greater than $\delta$ to ensure the reliability of the predicted depth. The loss is defined as follows. For simplicity, we show the calculation for one pixel $p$.
\begin{equation}
    L_{render} = \sum_{p\in I}\lambda_1\left\|\tilde{c}(p) - I(p)\right\|_{2}^{2} + \lambda_2\mathbb{I} (U(p) > \delta)\left\| \tilde{d}(p) - D(p) \right\|,
\end{equation}
where $\mathbb{I}$ is the indicator function, $\lambda_i$ sets the influence of each loss. The final objective from each level can be constructed as follows:
\begin{equation}
    L = \sum_{l=1}^3 L_{render}^l + \lambda_3 L_{mvs}^l.
\end{equation}

\section{Experiments} \label{sec:experiments}

In this section, we first describe the details including the datasets and implementation details, then we show the qualitative and quantitative comparisons of our network with state-of-the-art methods. Finally, we perform ablation studies to validate the effectiveness of our proposed method.

\subsection{Dataset and network details}

\paragraph{\textbf{Dataset.}}  
We train our framework end-to-end on the DTU~\cite{aanaes2016large} dataset to learn a generalizable network, using the same training and testing splits as MVSNeRF~\cite{chen2021mvsnerf}.
The DTU dataset is divided into 88 training scenes and 16 testing scenes with an image resolution of $512 \times 640$.
We evaluate the generalization ability of our method on the Realistic Synthetic NeRF \cite{mildenhall2021nerf} data and the Forward-Facing data~\cite{mildenhall2019local} by using the model merely trained on the DTU dataset. They both include 8 complex scenes that have a different distribution from DTU.

\paragraph{\textbf{Network details.}} We set $N=3$ for nearby source images, and implement the image feature extraction network using a FPN \cite{lin2017feature} like architecture. 
The number of depth hypotheses $[N_r^l]$ for encoding volume and sampling points $[N_d^l]$ for radiance field reconstruction at three stages is set to [48, 32, 8] and [8, 8, 4], respectively. 
And the number of adaptive neighbors $K$ is set to 8.
For the weight of each loss $\lambda_i$, $\lambda_1$ and $\lambda_3$ are set same value of 1.0, and $\lambda_2$ the weight of the depth loss grows linearly based on the training steps from $1e^{-4}$ to $1e^{-2}$.
The MLP decoder in rendering is similar to the MVSNeRF \cite{chen2021mvsnerf}.

We train our network using four V100 GPUs with a batch size of 1.
During training, 2048 pixels are randomly sampled from one novel viewpoint.
The model is trained with the Adam optimizer for 8 epochs with a base learning of $5e^{-4}$.
The learning rate is halved iteratively at the $2$-th, $4$-th, and $8$-th epoch.

\begin{table}[t]
  % \small
  \centering
  \caption{Quantitative results of reconstructed depth. We evaluate our depths and points reconstruction on the DTU and compare it with other methods. Our method significantly outperforms other neural rendering methods.}
  \begin{tabular}{
p{1.7cm}
p{1.6cm}<{\centering}p{1.6cm}<{\centering}p{1.8cm}<{\centering}
}
    \toprule
    \multicolumn{4}{c}{Novel-view depth metric} \\
    \midrule
    Method & Abs err $\downarrow$ & Acc(2mm) $\uparrow$ & Acc(10mm) $\uparrow$ \\
    \midrule
    PixelNeRF  & 47.8 & 0.039 & 0.187 \\
    IBRNet     & 324  & 0.000 & 0.866 \\
    MVSNeRF    & 7.00 & 0.717 & 0.866 \\
    ENeRF      & 4.60 & 0.792 & 0.917 \\
    Ours       & \textbf{4.29} & \textbf{0.867} & \textbf{0.937} \\
    \midrule
    \multicolumn{4}{c}{Reconstructed points metric} \\
    \midrule
    Method & Overall $\downarrow$ & Acc $\downarrow$ & Comp $\downarrow$ \\
    \midrule
    MVSNeRF    & 0.588 & 0.641 & 0.534 \\
    ENeRF      & 2.055 & 2.354 & 1.755 \\
    Ours       & \textbf{0.368} & \textbf{0.355} & \textbf{0.380} \\
    \bottomrule

\end{tabular}
  \label{tab:res-dtu-depth}
\end{table}

\begin{figure*}[t]
  \centering
  \includegraphics[width=1.0\textwidth]{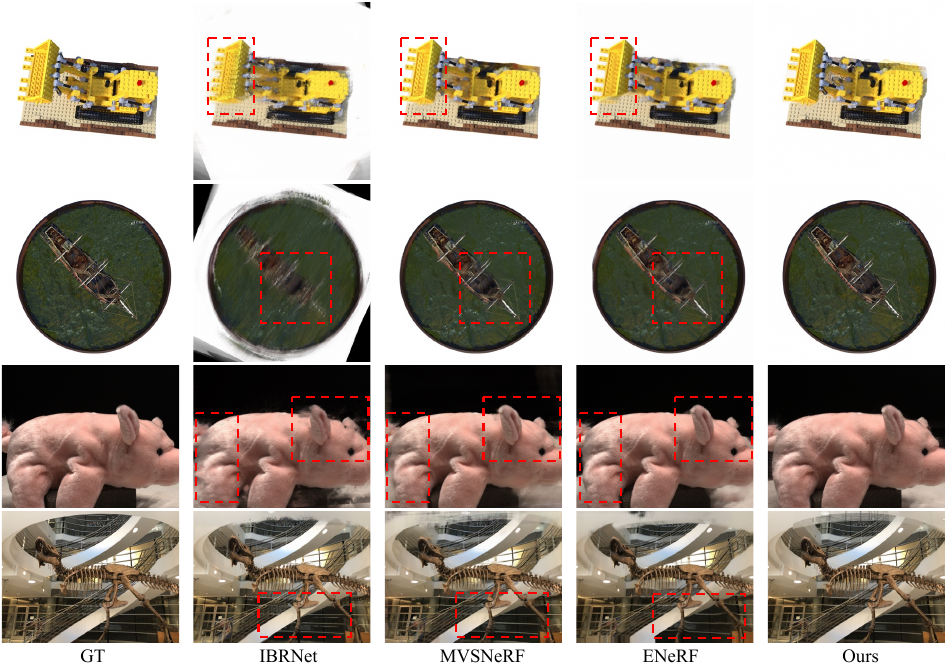}
  \caption{Qualitative comparison of rendering quality on diverse scenes between our method and state-of-the-art counterparts.We show the results of all three datasets. Our method renders better textures than other methods.
  This is particularly evident in the scenes highlighted in the red dashed boxes. (Best viewed by zooming in).}
  \label{fig:res-color}
\end{figure*}

\subsection{Evaluation results}
To evaluate our model, we compare it against current top-performing techniques for view synthesis, detailed below. We adopt the PSNR, SSIM, and LPIPS as the quantitative results for the rendered images. We also evaluate the geometry reconstruction quality by comparing depth and points reconstruction results on the DTU dataset.

\paragraph{\textbf{Quality of rendered image.}} Tab. \ref{tab:res_color} shows the comparison results with recent concurrent works with PSNR, SSIM, and LPIPS; 
For Realistic Synthetic NeRF~\cite{mildenhall2021nerf}, objects are primarily located in the central area, with the surrounding regions consisting of a white background. Existing methods typically utilize cropping to evaluate only the central portion of the image, while some methods evaluate the entire image region. In this paper, we explore two different approaches.
Our results on the DTU are significantly better than the comparison methods, where also 3 views are provided for the source images.
To show that our method has good generalizability, we also evaluate our model on the NeRF Synthetic and Real Forward-Facing datasets, following the experiment setting in MVSNeRF. For the NeRF Synthetic dataset, the quality of our results is further boosted significantly, leading to the best PSNR, SSIM, and LIPIPS in all compared methods. And also generates results that are comparable to ENeRF on the Real Forward-Facing dataset. As shown in Fig.~\ref{fig:res-color}, our results on these scenes are of very high visual quality compared with other methods.

\paragraph{\textbf{Quality of reconstructed depth.}}
To improve the quality of geometry, our approach reconstructs a density field by sharing the encoding volume. We evaluate our geometry reconstruction quality by comparing depth reconstruction and points reconstruction results. To reconstruct the final point, we follow ~\cite{yao2018mvsnet} to fuse the depth from multiple views. We compare our approach with the recent multi-view based methods on the DTU testing set.

As shown in Tab.~\ref{tab:res-dtu-depth}, benefit to our shared encoding volume, our approach achieves significantly more accurate depth rendering than others. Especially for 2mm accuracy, the metric is improved by \textbf{9.3\%} compared with ENeRF. For the quantitative evaluation of point cloud, we calculate the accuracy and completeness by the MATLAB code provided by the DTU dataset. We can see that our method outperforms other methods in both completeness and accuracy quality and rank the first place.

\begin{figure}[t]
  \centering
  \includegraphics[width=0.50\textwidth]{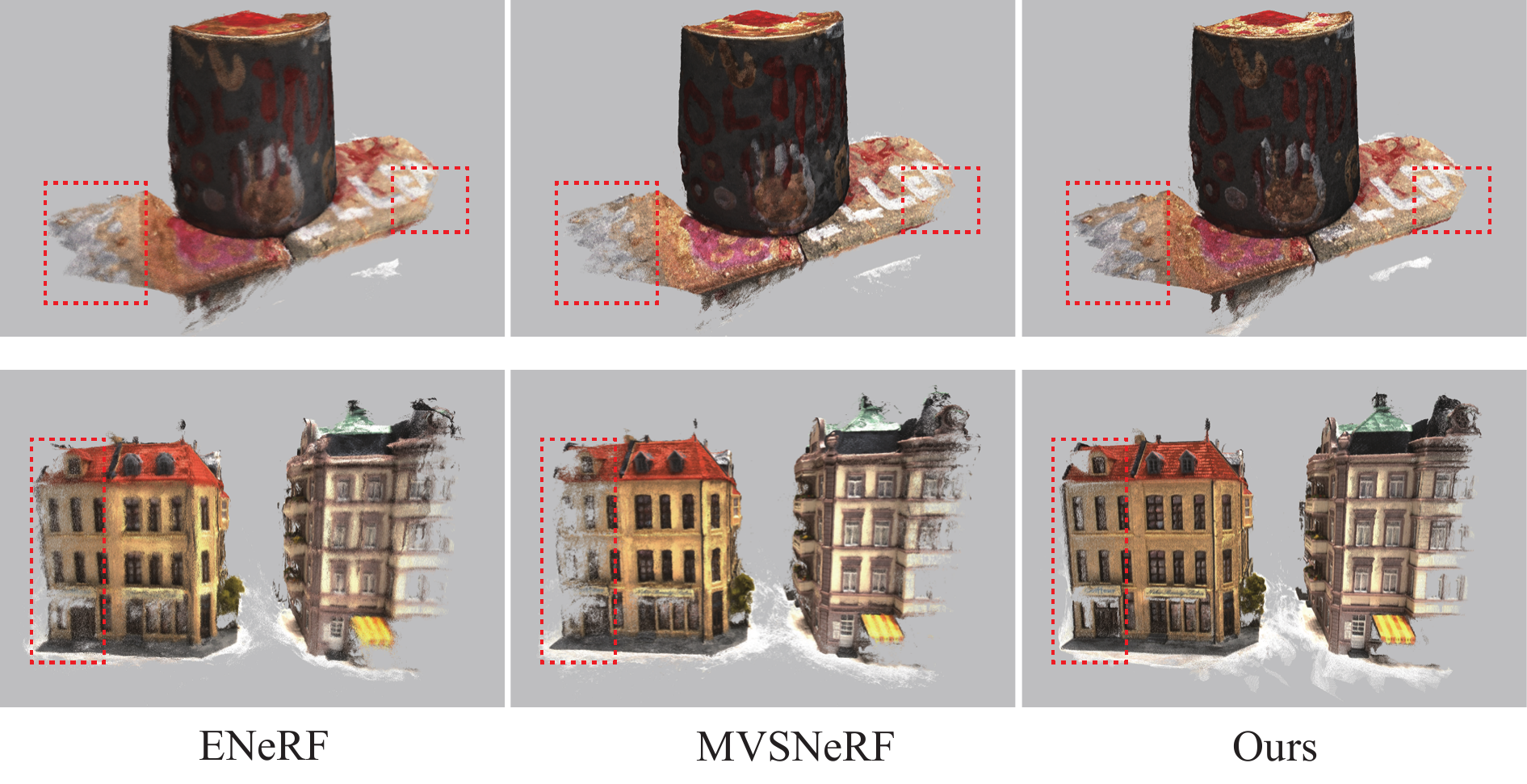}
  \caption{Qualitative comparisons between other rendering methods. we follow ~\cite{yao2018mvsnet} to fuse the depth from multiple views and reconstruct points. Our method achieves increasingly dense reconstruction. }
  \label{fig:res-points}
\end{figure}

The qualitative results of depths and points are shown in Fig. ~\ref{fig:1_depth_img} and Fig.~\ref{fig:res-points}. For the qualitative depths results, our method renders more accurate depth maps and includes more texture. Thanks to the accurate depths, ours generates more complete point clouds with finer details.

\subsection{Ablations and analysis}

To further demonstrate the contribution of our method, we introduce some groups of ablation studies on the DTU dataset in this section.

\paragraph{\textbf{Contribution of each proposed module.}}
In this section, we conduct experiments to verify the effectiveness of our proposed neural radiance field reconstruction.
As shown in Tab.~\ref{tbl:multi_scale}, when reconstructing the density field using the 3D position and viewing direction of the point, not shared volume, the rendering quality drops a lot. Besides, we remove the feature fusion module for comparison, this module shows the same good performance. Finally, we evaluate the influence of the pseudo-depth on photo-realistic quantity. As shown in the $3$-th row, using pseudo-depth supervision also improves the rendering performance.

\begin{figure}[t]
  \centering
  \includegraphics[width=0.50\textwidth]{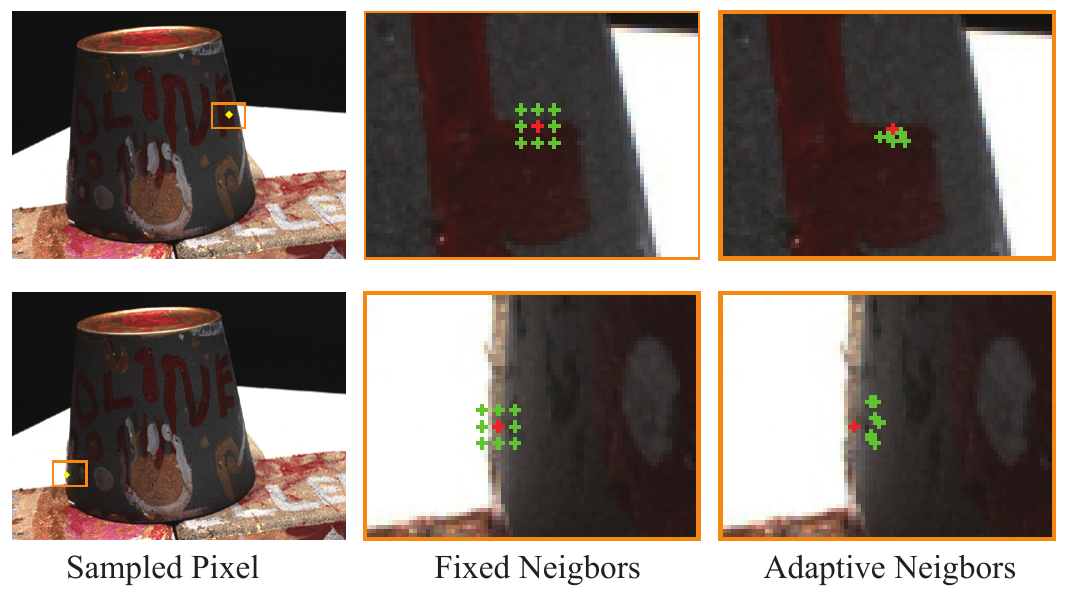}
  \caption{Visualization of adaptively sampled neighbors. The left column shows the center pixels, the middle column shows the eight fixed neighbors and the right column shows the learned adaptive locations.}
  \label{fig:ablation-neigs}
\end{figure}

\paragraph{\textbf{Contribution of depth-guided adaptive feature fusion.}} 
To verify the effectiveness of the proposed depth-guided adaptive feature fusion, we conduct experiments comparing with no-neighbor and different numbers of neighbors setting.
As shown in Tab.~\ref{tbl:ablation_fusion_adaptive}, our adaptive method performs the best in various settings of the number of neighbors.
As shown in Fig. \ref{fig:ablation-neigs}, the sampled neighbors tend to the locations which have salient features and have the same texture as the center pixel.

\paragraph{\textbf{Ablation on pseudo-depth supervised.}}
The auxiliary depth prediction head predicts a more accurate depth map compared to the regression approach, described in Sec.~\ref{sec:aux-depth-head}. The pseudo-depth supervising can help to reconstruct a more effective density field, and the photo-realistic quantity is shown in Tab. ~\ref{tbl:multi_scale}. To evaluate the geometry quaintly, we set five experiments in Tab ~\ref{tbl:abl_end2end}. The results show that more accurate pixels can benefit to improve the geometry quaintly.

\begin{table}[t]
\centering
\caption{Ablation studies for our proposed neural radiance field reconstruction of the photo-realistic quantity on DTU evaluation set.
}
\begin{tabular}{
m{4cm}
m{1.cm}<{\centering}m{1.cm}<{\centering}m{1.cm}<{\centering}
}
\toprule
Method & PSNR $\uparrow$ & SSIM $\uparrow$ & LPIPS $\downarrow$ \\
\midrule
w/o shared volume & 28.16 & 0.949 & 0.113 \\
w/o feature fusion & 28.26 & 0.950 & 0.110 \\
w/o pseudo-depth supervised & 28.08 & 0.948 & 0.115 \\
Ours & 28.52 & 0.952 & 0.110 \\
\bottomrule
\end{tabular}
\label{tbl:multi_scale}
\end{table}

\begin{table}[t]
  \centering
  \caption{Ablation experiments for depth-guided adaptive feature fusion.
  }
  \begin{tabular}{
m{2.0cm}
m{1.7cm}<{\centering}m{1.7cm}<{\centering}m{1.7cm}<{\centering}
}
\toprule
Method & PSNR $\uparrow$ & SSIM $\uparrow$ & LPIPS $\downarrow$ \\
\midrule
w/o neigs & 28.26 & 0.950 & 0.110 \\
4-neigs & 28.38 & 0.951 & 0.109 \\
8-neigs \ (Ours) & 28.53 & 0.952 & 0.108 \\
\bottomrule
\end{tabular}
  \label{tbl:ablation_fusion_adaptive}
\end{table}

\begin{table}[t]
\centering
\caption{Ablation experiments for pseudo-depth supervision of the geometry quantity on the DTU evaluation set. $\delta = 0.0$ means all pixels are used to supervise the rendered depths, $\delta = 1.0$ means not using the pseudo-depth supervised, and value between [0, 1] means using the pixels with a confidence score greater than $\delta$.}
\begin{tabular}{
m{2.cm}
m{1.7cm}<{\centering}m{1.7cm}<{\centering}m{1.7cm}<{\centering}
}
\toprule
Setting &Abs err $\downarrow$ & Acc(2mm) $\uparrow$ & Acc(10mm) $\uparrow$ \\
\midrule
$\delta = 0.0$ & 4.42 & 0.852 & 0.925 \\
$\delta = 0.3$ & 4.32 & 0.867 & 0.935 \\
$\delta = 0.5$\ (Ours) & 4.29 & 0.867 & 0.937 \\
$\delta = 0.7$ & 4.33 & 0.858 & 0.937 \\
$\delta = 1.0$ & 5.36 & 0.786 & 0.920 \\
\bottomrule
\end{tabular}
\label{tbl:abl_end2end}
\end{table}

\section{Conclusion}
This work proposes an end-to-end framework for generalizable radiance field reconstruction.
In this framework, an auxiliary depth prediction head is additionally learned to provide depth priors when performing point sampling and pyramid encoding volumes construction.
The constructed pyramid volumes provide multi-scale geometric information of the scene, thereby improving rendering performance across scenes.
To mitigate the effects of factors such as occlusion and illumination on the constructed encoding volumes, we enhance the volume features by depth-guided adaptive feature fusion.
Our method generalizes well across diverse testing datasets and can significantly outperform concurrent works on photo-realistic and geometry of rendered novel views.

\bibliographystyle{ACM-Reference-Format}
\bibliography{main}


\begin{thebibliography}{36}

%%% the Web address.

\ifx \showCODEN    \undefined \def \showCODEN     #1{\unskip}     \fi
\ifx \showDOI      \undefined \def \showDOI       #1{#1}\fi
\ifx \showISBNx    \undefined \def \showISBNx     #1{\unskip}     \fi
\ifx \showISBNxiii \undefined \def \showISBNxiii  #1{\unskip}     \fi
\ifx \showISSN     \undefined \def \showISSN      #1{\unskip}     \fi
\ifx \showLCCN     \undefined \def \showLCCN      #1{\unskip}     \fi
\ifx \shownote     \undefined \def \shownote      #1{#1}          \fi
\ifx \showarticletitle \undefined \def \showarticletitle #1{#1}   \fi
\ifx \showURL      \undefined \def \showURL       {\relax}        \fi
% invisible to TeX
\providecommand\bibfield[2]{#2}
\providecommand\bibinfo[2]{#2}
\providecommand\natexlab[1]{#1}
\providecommand\showeprint[2][]{arXiv:#2}

\bibitem[Aan{\ae}s et~al\mbox{.}(2016)]%
        {aanaes2016large}
\bibfield{author}{\bibinfo{person}{Henrik Aan{\ae}s},
  \bibinfo{person}{Rasmus~Ramsb{\o}l Jensen}, \bibinfo{person}{George
  Vogiatzis}, \bibinfo{person}{Engin Tola}, {and}
  \bibinfo{person}{Anders~Bjorholm Dahl}.} \bibinfo{year}{2016}\natexlab{}.
\newblock \showarticletitle{Large-scale data for multiple-view stereopsis}.
\newblock \bibinfo{journal}{\emph{International Journal of Computer Vision}}
  \bibinfo{volume}{120}, \bibinfo{number}{2} (\bibinfo{year}{2016}),
  \bibinfo{pages}{153--168}.
\newblock


\bibitem[Bi et~al\mbox{.}(2020a)]%
        {bi2020neural}
\bibfield{author}{\bibinfo{person}{Sai Bi}, \bibinfo{person}{Zexiang Xu},
  \bibinfo{person}{Pratul Srinivasan}, \bibinfo{person}{Ben Mildenhall},
  \bibinfo{person}{Kalyan Sunkavalli}, \bibinfo{person}{Milo{\v{s}}
  Ha{\v{s}}an}, \bibinfo{person}{Yannick Hold-Geoffroy}, \bibinfo{person}{David
  Kriegman}, {and} \bibinfo{person}{Ravi Ramamoorthi}.}
  \bibinfo{year}{2020}\natexlab{a}.
\newblock \showarticletitle{Neural reflectance fields for appearance
  acquisition}.
\newblock \bibinfo{journal}{\emph{arXiv preprint arXiv:2008.03824}}
  (\bibinfo{year}{2020}).
\newblock


\bibitem[Bi et~al\mbox{.}(2020b)]%
        {bi2020deep}
\bibfield{author}{\bibinfo{person}{Sai Bi}, \bibinfo{person}{Zexiang Xu},
  \bibinfo{person}{Kalyan Sunkavalli}, \bibinfo{person}{Milo{\v{s}}
  Ha{\v{s}}an}, \bibinfo{person}{Yannick Hold-Geoffroy}, \bibinfo{person}{David
  Kriegman}, {and} \bibinfo{person}{Ravi Ramamoorthi}.}
  \bibinfo{year}{2020}\natexlab{b}.
\newblock \showarticletitle{Deep reflectance volumes: Relightable
  reconstructions from multi-view photometric images}. In
  \bibinfo{booktitle}{\emph{European Conference on Computer Vision}}. Springer,
  \bibinfo{pages}{294--311}.
\newblock


\bibitem[Boss et~al\mbox{.}(2021)]%
        {boss2021nerd}
\bibfield{author}{\bibinfo{person}{Mark Boss}, \bibinfo{person}{Raphael Braun},
  \bibinfo{person}{Varun Jampani}, \bibinfo{person}{Jonathan~T Barron},
  \bibinfo{person}{Ce Liu}, {and} \bibinfo{person}{Hendrik Lensch}.}
  \bibinfo{year}{2021}\natexlab{}.
\newblock \showarticletitle{Nerd: Neural reflectance decomposition from image
  collections}. In \bibinfo{booktitle}{\emph{Proceedings of the IEEE/CVF
  International Conference on Computer Vision}}. \bibinfo{pages}{12684--12694}.
\newblock


\bibitem[Chen et~al\mbox{.}(2021)]%
        {chen2021mvsnerf}
\bibfield{author}{\bibinfo{person}{Anpei Chen}, \bibinfo{person}{Zexiang Xu},
  \bibinfo{person}{Fuqiang Zhao}, \bibinfo{person}{Xiaoshuai Zhang},
  \bibinfo{person}{Fanbo Xiang}, \bibinfo{person}{Jingyi Yu}, {and}
  \bibinfo{person}{Hao Su}.} \bibinfo{year}{2021}\natexlab{}.
\newblock \showarticletitle{Mvsnerf: Fast generalizable radiance field
  reconstruction from multi-view stereo}. In
  \bibinfo{booktitle}{\emph{Proceedings of the IEEE/CVF International
  Conference on Computer Vision}}. \bibinfo{pages}{14124--14133}.
\newblock


\bibitem[Deng et~al\mbox{.}(2022)]%
        {deng2022depth}
\bibfield{author}{\bibinfo{person}{Kangle Deng}, \bibinfo{person}{Andrew Liu},
  \bibinfo{person}{Jun-Yan Zhu}, {and} \bibinfo{person}{Deva Ramanan}.}
  \bibinfo{year}{2022}\natexlab{}.
\newblock \showarticletitle{Depth-supervised nerf: Fewer views and faster
  training for free}. In \bibinfo{booktitle}{\emph{Proceedings of the IEEE/CVF
  Conference on Computer Vision and Pattern Recognition}}.
  \bibinfo{pages}{12882--12891}.
\newblock


\bibitem[Gu et~al\mbox{.}(2020)]%
        {gu2020cascade}
\bibfield{author}{\bibinfo{person}{Xiaodong Gu}, \bibinfo{person}{Zhiwen Fan},
  \bibinfo{person}{Siyu Zhu}, \bibinfo{person}{Zuozhuo Dai},
  \bibinfo{person}{Feitong Tan}, {and} \bibinfo{person}{Ping Tan}.}
  \bibinfo{year}{2020}\natexlab{}.
\newblock \showarticletitle{Cascade cost volume for high-resolution multi-view
  stereo and stereo matching}. In \bibinfo{booktitle}{\emph{Proceedings of the
  IEEE/CVF Conference on Computer Vision and Pattern Recognition}}.
  \bibinfo{pages}{2495--2504}.
\newblock


\bibitem[Kania et~al\mbox{.}(2022)]%
        {kania2022conerf}
\bibfield{author}{\bibinfo{person}{Kacper Kania}, \bibinfo{person}{Kwang~Moo
  Yi}, \bibinfo{person}{Marek Kowalski}, \bibinfo{person}{Tomasz
  Trzci{\'n}ski}, {and} \bibinfo{person}{Andrea Tagliasacchi}.}
  \bibinfo{year}{2022}\natexlab{}.
\newblock \showarticletitle{Conerf: Controllable neural radiance fields}. In
  \bibinfo{booktitle}{\emph{Proceedings of the IEEE/CVF Conference on Computer
  Vision and Pattern Recognition}}. \bibinfo{pages}{18623--18632}.
\newblock


\bibitem[Khot et~al\mbox{.}(2019)]%
        {khot2019learning}
\bibfield{author}{\bibinfo{person}{Tejas Khot}, \bibinfo{person}{Shubham
  Agrawal}, \bibinfo{person}{Shubham Tulsiani}, \bibinfo{person}{Christoph
  Mertz}, \bibinfo{person}{Simon Lucey}, {and} \bibinfo{person}{Martial
  Hebert}.} \bibinfo{year}{2019}\natexlab{}.
\newblock \showarticletitle{Learning unsupervised multi-view stereopsis via
  robust photometric consistency}.
\newblock \bibinfo{journal}{\emph{arXiv preprint arXiv:1905.02706}}
  (\bibinfo{year}{2019}).
\newblock


\bibitem[Lin et~al\mbox{.}(2022)]%
        {lin2022efficient}
\bibfield{author}{\bibinfo{person}{Haotong Lin}, \bibinfo{person}{Sida Peng},
  \bibinfo{person}{Zhen Xu}, \bibinfo{person}{Yunzhi Yan},
  \bibinfo{person}{Qing Shuai}, \bibinfo{person}{Hujun Bao}, {and}
  \bibinfo{person}{Xiaowei Zhou}.} \bibinfo{year}{2022}\natexlab{}.
\newblock \showarticletitle{Efficient Neural Radiance Fields for Interactive
  Free-viewpoint Video}. In \bibinfo{booktitle}{\emph{SIGGRAPH Asia 2022
  Conference Papers}}. \bibinfo{pages}{1--9}.
\newblock


\bibitem[Lin et~al\mbox{.}(2017)]%
        {lin2017feature}
\bibfield{author}{\bibinfo{person}{Tsung-Yi Lin}, \bibinfo{person}{Piotr
  Doll{\'a}r}, \bibinfo{person}{Ross Girshick}, \bibinfo{person}{Kaiming He},
  \bibinfo{person}{Bharath Hariharan}, {and} \bibinfo{person}{Serge Belongie}.}
  \bibinfo{year}{2017}\natexlab{}.
\newblock \showarticletitle{Feature pyramid networks for object detection}. In
  \bibinfo{booktitle}{\emph{Proceedings of the IEEE conference on computer
  vision and pattern recognition}}. \bibinfo{pages}{2117--2125}.
\newblock


\bibitem[Liu et~al\mbox{.}(2020)]%
        {liu2020neural}
\bibfield{author}{\bibinfo{person}{Lingjie Liu}, \bibinfo{person}{Jiatao Gu},
  \bibinfo{person}{Kyaw Zaw~Lin}, \bibinfo{person}{Tat-Seng Chua}, {and}
  \bibinfo{person}{Christian Theobalt}.} \bibinfo{year}{2020}\natexlab{}.
\newblock \showarticletitle{Neural sparse voxel fields}.
\newblock \bibinfo{journal}{\emph{Advances in Neural Information Processing
  Systems}}  \bibinfo{volume}{33} (\bibinfo{year}{2020}),
  \bibinfo{pages}{15651--15663}.
\newblock


\bibitem[Lombardi et~al\mbox{.}(2019)]%
        {lombardi2019neural}
\bibfield{author}{\bibinfo{person}{Stephen Lombardi}, \bibinfo{person}{Tomas
  Simon}, \bibinfo{person}{Jason Saragih}, \bibinfo{person}{Gabriel Schwartz},
  \bibinfo{person}{Andreas Lehrmann}, {and} \bibinfo{person}{Yaser Sheikh}.}
  \bibinfo{year}{2019}\natexlab{}.
\newblock \showarticletitle{Neural volumes: Learning dynamic renderable volumes
  from images}.
\newblock \bibinfo{journal}{\emph{arXiv preprint arXiv:1906.07751}}
  (\bibinfo{year}{2019}).
\newblock


\bibitem[Luo et~al\mbox{.}(2019)]%
        {luo2019p}
\bibfield{author}{\bibinfo{person}{Keyang Luo}, \bibinfo{person}{Tao Guan},
  \bibinfo{person}{Lili Ju}, \bibinfo{person}{Haipeng Huang}, {and}
  \bibinfo{person}{Yawei Luo}.} \bibinfo{year}{2019}\natexlab{}.
\newblock \showarticletitle{P-mvsnet: Learning patch-wise matching confidence
  aggregation for multi-view stereo}. In \bibinfo{booktitle}{\emph{Proceedings
  of the IEEE/CVF International Conference on Computer Vision}}.
  \bibinfo{pages}{10452--10461}.
\newblock


\bibitem[Martin-Brualla et~al\mbox{.}(2021)]%
        {martin2021nerfw}
\bibfield{author}{\bibinfo{person}{Ricardo Martin-Brualla},
  \bibinfo{person}{Noha Radwan}, \bibinfo{person}{Mehdi~SM Sajjadi},
  \bibinfo{person}{Jonathan~T Barron}, \bibinfo{person}{Alexey Dosovitskiy},
  {and} \bibinfo{person}{Daniel Duckworth}.} \bibinfo{year}{2021}\natexlab{}.
\newblock \showarticletitle{Nerf in the wild: Neural radiance fields for
  unconstrained photo collections}. In \bibinfo{booktitle}{\emph{Proceedings of
  the IEEE/CVF Conference on Computer Vision and Pattern Recognition}}.
  \bibinfo{pages}{7210--7219}.
\newblock


\bibitem[Mildenhall et~al\mbox{.}(2019)]%
        {mildenhall2019local}
\bibfield{author}{\bibinfo{person}{Ben Mildenhall}, \bibinfo{person}{Pratul~P
  Srinivasan}, \bibinfo{person}{Rodrigo Ortiz-Cayon},
  \bibinfo{person}{Nima~Khademi Kalantari}, \bibinfo{person}{Ravi Ramamoorthi},
  \bibinfo{person}{Ren Ng}, {and} \bibinfo{person}{Abhishek Kar}.}
  \bibinfo{year}{2019}\natexlab{}.
\newblock \showarticletitle{Local light field fusion: Practical view synthesis
  with prescriptive sampling guidelines}.
\newblock \bibinfo{journal}{\emph{ACM Transactions on Graphics (TOG)}}
  \bibinfo{volume}{38}, \bibinfo{number}{4} (\bibinfo{year}{2019}),
  \bibinfo{pages}{1--14}.
\newblock


\bibitem[Mildenhall et~al\mbox{.}(2021)]%
        {mildenhall2021nerf}
\bibfield{author}{\bibinfo{person}{Ben Mildenhall}, \bibinfo{person}{Pratul~P
  Srinivasan}, \bibinfo{person}{Matthew Tancik}, \bibinfo{person}{Jonathan~T
  Barron}, \bibinfo{person}{Ravi Ramamoorthi}, {and} \bibinfo{person}{Ren Ng}.}
  \bibinfo{year}{2021}\natexlab{}.
\newblock \showarticletitle{Nerf: Representing scenes as neural radiance fields
  for view synthesis}.
\newblock \bibinfo{journal}{\emph{Commun. ACM}} \bibinfo{volume}{65},
  \bibinfo{number}{1} (\bibinfo{year}{2021}), \bibinfo{pages}{99--106}.
\newblock


\bibitem[Roessle et~al\mbox{.}(2022)]%
        {roessle2022dense}
\bibfield{author}{\bibinfo{person}{Barbara Roessle},
  \bibinfo{person}{Jonathan~T Barron}, \bibinfo{person}{Ben Mildenhall},
  \bibinfo{person}{Pratul~P Srinivasan}, {and} \bibinfo{person}{Matthias
  Nie{\ss}ner}.} \bibinfo{year}{2022}\natexlab{}.
\newblock \showarticletitle{Dense depth priors for neural radiance fields from
  sparse input views}. In \bibinfo{booktitle}{\emph{Proceedings of the IEEE/CVF
  Conference on Computer Vision and Pattern Recognition}}.
  \bibinfo{pages}{12892--12901}.
\newblock


\bibitem[Sch\"{o}nberger and Frahm(2016)]%
        {schoenberger2016sfm}
\bibfield{author}{\bibinfo{person}{Johannes~Lutz Sch\"{o}nberger} {and}
  \bibinfo{person}{Jan-Michael Frahm}.} \bibinfo{year}{2016}\natexlab{}.
\newblock \showarticletitle{Structure-from-Motion Revisited}. In
  \bibinfo{booktitle}{\emph{Conference on Computer Vision and Pattern
  Recognition (CVPR)}}.
\newblock


\bibitem[Sitzmann et~al\mbox{.}(2019)]%
        {sitzmann2019deepvoxels}
\bibfield{author}{\bibinfo{person}{Vincent Sitzmann}, \bibinfo{person}{Justus
  Thies}, \bibinfo{person}{Felix Heide}, \bibinfo{person}{Matthias
  Nie{\ss}ner}, \bibinfo{person}{Gordon Wetzstein}, {and}
  \bibinfo{person}{Michael Zollhofer}.} \bibinfo{year}{2019}\natexlab{}.
\newblock \showarticletitle{Deepvoxels: Learning persistent 3d feature
  embeddings}. In \bibinfo{booktitle}{\emph{Proceedings of the IEEE/CVF
  Conference on Computer Vision and Pattern Recognition}}.
  \bibinfo{pages}{2437--2446}.
\newblock


\bibitem[Srinivasan et~al\mbox{.}(2021)]%
        {srinivasan2021nerv}
\bibfield{author}{\bibinfo{person}{Pratul~P Srinivasan},
  \bibinfo{person}{Boyang Deng}, \bibinfo{person}{Xiuming Zhang},
  \bibinfo{person}{Matthew Tancik}, \bibinfo{person}{Ben Mildenhall}, {and}
  \bibinfo{person}{Jonathan~T Barron}.} \bibinfo{year}{2021}\natexlab{}.
\newblock \showarticletitle{Nerv: Neural reflectance and visibility fields for
  relighting and view synthesis}. In \bibinfo{booktitle}{\emph{Proceedings of
  the IEEE/CVF Conference on Computer Vision and Pattern Recognition}}.
  \bibinfo{pages}{7495--7504}.
\newblock


\bibitem[Tancik et~al\mbox{.}(2022)]%
        {tancik2022block}
\bibfield{author}{\bibinfo{person}{Matthew Tancik}, \bibinfo{person}{Vincent
  Casser}, \bibinfo{person}{Xinchen Yan}, \bibinfo{person}{Sabeek Pradhan},
  \bibinfo{person}{Ben Mildenhall}, \bibinfo{person}{Pratul~P Srinivasan},
  \bibinfo{person}{Jonathan~T Barron}, {and} \bibinfo{person}{Henrik
  Kretzschmar}.} \bibinfo{year}{2022}\natexlab{}.
\newblock \showarticletitle{Block-nerf: Scalable large scene neural view
  synthesis}. In \bibinfo{booktitle}{\emph{Proceedings of the IEEE/CVF
  Conference on Computer Vision and Pattern Recognition}}.
  \bibinfo{pages}{8248--8258}.
\newblock


\bibitem[Turki et~al\mbox{.}(2022)]%
        {turki2022mega}
\bibfield{author}{\bibinfo{person}{Haithem Turki}, \bibinfo{person}{Deva
  Ramanan}, {and} \bibinfo{person}{Mahadev Satyanarayanan}.}
  \bibinfo{year}{2022}\natexlab{}.
\newblock \showarticletitle{Mega-NeRF: Scalable Construction of Large-Scale
  NeRFs for Virtual Fly-Throughs}. In \bibinfo{booktitle}{\emph{Proceedings of
  the IEEE/CVF Conference on Computer Vision and Pattern Recognition}}.
  \bibinfo{pages}{12922--12931}.
\newblock


\bibitem[Wang et~al\mbox{.}(2021a)]%
        {wang2021patchmatchnet}
\bibfield{author}{\bibinfo{person}{Fangjinhua Wang}, \bibinfo{person}{Silvano
  Galliani}, \bibinfo{person}{Christoph Vogel}, \bibinfo{person}{Pablo
  Speciale}, {and} \bibinfo{person}{Marc Pollefeys}.}
  \bibinfo{year}{2021}\natexlab{a}.
\newblock \showarticletitle{Patchmatchnet: Learned multi-view patchmatch
  stereo}. In \bibinfo{booktitle}{\emph{Proceedings of the IEEE/CVF Conference
  on Computer Vision and Pattern Recognition}}. \bibinfo{pages}{14194--14203}.
\newblock


\bibitem[Wang et~al\mbox{.}(2021b)]%
        {wang2021ibrnet}
\bibfield{author}{\bibinfo{person}{Qianqian Wang}, \bibinfo{person}{Zhicheng
  Wang}, \bibinfo{person}{Kyle Genova}, \bibinfo{person}{Pratul~P Srinivasan},
  \bibinfo{person}{Howard Zhou}, \bibinfo{person}{Jonathan~T Barron},
  \bibinfo{person}{Ricardo Martin-Brualla}, \bibinfo{person}{Noah Snavely},
  {and} \bibinfo{person}{Thomas Funkhouser}.} \bibinfo{year}{2021}\natexlab{b}.
\newblock \showarticletitle{Ibrnet: Learning multi-view image-based rendering}.
  In \bibinfo{booktitle}{\emph{Proceedings of the IEEE/CVF Conference on
  Computer Vision and Pattern Recognition}}. \bibinfo{pages}{4690--4699}.
\newblock


\bibitem[Wu et~al\mbox{.}(2022)]%
        {wu2022object}
\bibfield{author}{\bibinfo{person}{Qianyi Wu}, \bibinfo{person}{Xian Liu},
  \bibinfo{person}{Yuedong Chen}, \bibinfo{person}{Kejie Li},
  \bibinfo{person}{Chuanxia Zheng}, \bibinfo{person}{Jianfei Cai}, {and}
  \bibinfo{person}{Jianmin Zheng}.} \bibinfo{year}{2022}\natexlab{}.
\newblock \showarticletitle{Object-compositional neural implicit surfaces}. In
  \bibinfo{booktitle}{\emph{European Conference on Computer Vision}}. Springer,
  \bibinfo{pages}{197--213}.
\newblock


\bibitem[Xu and Zhang(2020)]%
        {xu2020aanet}
\bibfield{author}{\bibinfo{person}{Haofei Xu} {and} \bibinfo{person}{Juyong
  Zhang}.} \bibinfo{year}{2020}\natexlab{}.
\newblock \showarticletitle{Aanet: Adaptive aggregation network for efficient
  stereo matching}. In \bibinfo{booktitle}{\emph{Proceedings of the IEEE/CVF
  Conference on Computer Vision and Pattern Recognition}}.
  \bibinfo{pages}{1959--1968}.
\newblock


\bibitem[Xu et~al\mbox{.}(2021)]%
        {xu2021jdacs}
\bibfield{author}{\bibinfo{person}{Hongbin Xu}, \bibinfo{person}{Zhipeng Zhou},
  \bibinfo{person}{Yu Qiao}, \bibinfo{person}{Wenxiong Kang}, {and}
  \bibinfo{person}{Qiuxia Wu}.} \bibinfo{year}{2021}\natexlab{}.
\newblock \showarticletitle{Self-supervised Multi-view Stereo via Effective
  Co-Segmentation and Data-Augmentation}. In
  \bibinfo{booktitle}{\emph{Proceedings of the AAAI Conference on Artificial
  Intelligence}}.
\newblock


\bibitem[Xu et~al\mbox{.}(2022)]%
        {xu2022point}
\bibfield{author}{\bibinfo{person}{Qiangeng Xu}, \bibinfo{person}{Zexiang Xu},
  \bibinfo{person}{Julien Philip}, \bibinfo{person}{Sai Bi},
  \bibinfo{person}{Zhixin Shu}, \bibinfo{person}{Kalyan Sunkavalli}, {and}
  \bibinfo{person}{Ulrich Neumann}.} \bibinfo{year}{2022}\natexlab{}.
\newblock \showarticletitle{Point-nerf: Point-based neural radiance fields}. In
  \bibinfo{booktitle}{\emph{Proceedings of the IEEE/CVF Conference on Computer
  Vision and Pattern Recognition}}. \bibinfo{pages}{5438--5448}.
\newblock


\bibitem[Xu and Harada(2022)]%
        {xu2022deforming}
\bibfield{author}{\bibinfo{person}{Tianhan Xu} {and} \bibinfo{person}{Tatsuya
  Harada}.} \bibinfo{year}{2022}\natexlab{}.
\newblock \showarticletitle{Deforming Radiance Fields with Cages}. In
  \bibinfo{booktitle}{\emph{European Conference on Computer Vision}}. Springer,
  \bibinfo{pages}{159--175}.
\newblock


\bibitem[Yao et~al\mbox{.}(2018)]%
        {yao2018mvsnet}
\bibfield{author}{\bibinfo{person}{Yao Yao}, \bibinfo{person}{Zixin Luo},
  \bibinfo{person}{Shiwei Li}, \bibinfo{person}{Tian Fang}, {and}
  \bibinfo{person}{Long Quan}.} \bibinfo{year}{2018}\natexlab{}.
\newblock \showarticletitle{Mvsnet: Depth inference for unstructured multi-view
  stereo}. In \bibinfo{booktitle}{\emph{Proceedings of the European conference
  on computer vision (ECCV)}}. \bibinfo{pages}{767--783}.
\newblock


\bibitem[Yao et~al\mbox{.}(2019)]%
        {yao2019recurrent}
\bibfield{author}{\bibinfo{person}{Yao Yao}, \bibinfo{person}{Zixin Luo},
  \bibinfo{person}{Shiwei Li}, \bibinfo{person}{Tianwei Shen},
  \bibinfo{person}{Tian Fang}, {and} \bibinfo{person}{Long Quan}.}
  \bibinfo{year}{2019}\natexlab{}.
\newblock \showarticletitle{Recurrent mvsnet for high-resolution multi-view
  stereo depth inference}. In \bibinfo{booktitle}{\emph{Proceedings of the
  IEEE/CVF conference on computer vision and pattern recognition}}.
  \bibinfo{pages}{5525--5534}.
\newblock


\bibitem[Yu et~al\mbox{.}(2021)]%
        {yu2021pixelnerf}
\bibfield{author}{\bibinfo{person}{Alex Yu}, \bibinfo{person}{Vickie Ye},
  \bibinfo{person}{Matthew Tancik}, {and} \bibinfo{person}{Angjoo Kanazawa}.}
  \bibinfo{year}{2021}\natexlab{}.
\newblock \showarticletitle{pixelnerf: Neural radiance fields from one or few
  images}. In \bibinfo{booktitle}{\emph{Proceedings of the IEEE/CVF Conference
  on Computer Vision and Pattern Recognition}}. \bibinfo{pages}{4578--4587}.
\newblock


\bibitem[Yuan et~al\mbox{.}(2022)]%
        {yuan2022nerf}
\bibfield{author}{\bibinfo{person}{Yu-Jie Yuan}, \bibinfo{person}{Yang-Tian
  Sun}, \bibinfo{person}{Yu-Kun Lai}, \bibinfo{person}{Yuewen Ma},
  \bibinfo{person}{Rongfei Jia}, {and} \bibinfo{person}{Lin Gao}.}
  \bibinfo{year}{2022}\natexlab{}.
\newblock \showarticletitle{NeRF-editing: geometry editing of neural radiance
  fields}. In \bibinfo{booktitle}{\emph{Proceedings of the IEEE/CVF Conference
  on Computer Vision and Pattern Recognition}}. \bibinfo{pages}{18353--18364}.
\newblock


\bibitem[Zhang et~al\mbox{.}(2020)]%
        {zhang2020nerf++}
\bibfield{author}{\bibinfo{person}{Kai Zhang}, \bibinfo{person}{Gernot
  Riegler}, \bibinfo{person}{Noah Snavely}, {and} \bibinfo{person}{Vladlen
  Koltun}.} \bibinfo{year}{2020}\natexlab{}.
\newblock \showarticletitle{Nerf++: Analyzing and improving neural radiance
  fields}.
\newblock \bibinfo{journal}{\emph{arXiv preprint arXiv:2010.07492}}
  (\bibinfo{year}{2020}).
\newblock


\bibitem[Zhou et~al\mbox{.}(2018)]%
        {zhou2018stereo}
\bibfield{author}{\bibinfo{person}{Tinghui Zhou}, \bibinfo{person}{Richard
  Tucker}, \bibinfo{person}{John Flynn}, \bibinfo{person}{Graham Fyffe}, {and}
  \bibinfo{person}{Noah Snavely}.} \bibinfo{year}{2018}\natexlab{}.
\newblock \showarticletitle{Stereo magnification: Learning view synthesis using
  multiplane images}.
\newblock \bibinfo{journal}{\emph{arXiv preprint arXiv:1805.09817}}
  (\bibinfo{year}{2018}).
\newblock


\end{thebibliography}

\end{document}